\documentclass[runningheads]{llncs}
\usepackage[T1]{fontenc}
\usepackage{graphicx}

\usepackage{tikz}
\usepackage{amsmath,amssymb} 
\usepackage{color}
\usepackage{tikz}
\usepackage{color, colortbl}

\usepackage[accsupp]{axessibility}  
\definecolor{Gray}{gray}{0.5}
\definecolor{Gray2}{gray}{0.9}
\definecolor{Gray3}{gray}{0.9}
\definecolor{Gray4}{gray}{0.6}

\newcommand{\std}[1]{\color{Gray}\small{$\pm$#1}}

\usepackage{multirow}      
\newcommand{\real}{\mathbb{R}}
\newcommand{\x}{\mathbf{x}}
\newcommand{\m}{\mathbf{m}}
\newcommand{\s}{\mathbf{s}}
\newcommand\norm[1]{\left\lVert#1\right\rVert}

\newcommand{\tick}{\checkmark}
\newcommand{\mycomment}[1]{}

\usepackage[super]{nth}
\usepackage{booktabs}
\usepackage{tablefootnote}

\usepackage{bbm}
\newcommand{\mc}[1]{\multicolumn{1}{c}{#1}}

\newcommand{\tbf}[1]{\textbf{#1}}

\usepackage[colorinlistoftodos,prependcaption,textsize=small]{todonotes}
\usepackage{amssymb}
\usepackage{pifont}
\newcommand{\un}[1]{\underline{#1}}

\usepackage{subfigure}
\usepackage{graphicx}

\usepackage{hyperref}

\newcommand{\comment}[1]{}

\usepackage[normalem]{ulem}

\begin{document}

\title{Masked Image Modelling\\ for retinal OCT understanding}

\author{Anonymous}

\author{Theodoros Pissas\inst{1} \and 
Pablo Márquez-Neila \inst{1}\and
Sebastian Wolf\inst{2} \and \\ Martin Zinkernagel\inst{2}  \and Raphael Sznitman \inst{1} }

\institute{University of Bern, Bern, Switzerland \and
Department of Ophthalmology, Inselspital, Bern, Switzerland\\
\email{theodoros.pissas@unibe.ch}}

\maketitle              

\begin{abstract}
This work explores the effectiveness of masked image modelling for learning representations of retinal OCT images. To this end, we leverage Masked Autoencoders (MAE), a simple and scalable method for self-supervised learning, to obtain a powerful and general representation for OCT images by training on $700$K OCT images from $41$K patients collected under real world clinical settings. We also provide the first extensive evaluation for a model of OCT on a challenging battery of $6$ downstream tasks. Our model achieves strong performance when fully finetuned but can also serve as a versatile frozen feature extractor for many tasks using lightweight adapters. Furthermore, we propose an extension of the MAE pretraining to fuse OCT with an auxiliary modality, namely, IR fundus images and learn a joint model for both. We demonstrate our approach improves performance on a multimodal downstream application. Our experiments utilize most publicly available OCT datasets, thus enabling future comparisons. Our code and model weights are publicly available \url{https://github.com/TheoPis/MIM_OCT}.
\end{abstract}

\section{Introduction}
\label{sec:intro}
The swift and precise diagnosis, treatment, and monitoring of retinal diseases are crucial in preventing vision impairment. However, the rising number of patients and limited availability of expert clinicians, especially in regions with scarce healthcare resources, pose significant challenges \cite{ai_ophthalmology}. Optical Coherence Tomography (OCT), which visualizes the structural characteristics of the retina, plays a pivotal role in diagnosing and managing numerous ophthalmic diseases such as Age-Related Macular Degeneration (AMD) \cite{oct_AMD,oct_AMD_2}, Diabetic Retinopathy (DR) \cite{oct_dr}, and Glaucoma \cite{oct_glaucoma}. 

Deep learning models have shown promise in matching human performance for diagnostic and retinal image analysis tasks \cite{defauw,ai_medical_imaging,kurmann2019expert}, and can accelerate clinical trials by providing automated biomarker quantification on imaging outcomes \cite{ai_amd_clinical_trials}. However, existing methods for automating OCT understanding, such as disease classification \cite{kermany}, progression \cite{tinc}, biomarker detection \cite{kurmann2019expert}, and layer or fluid segmentation \cite{retouch,layer_seg_sanchez,layers_apostolopoulos}, are designed for narrow applications and trained on data from a relatively small number of patients.

Instead, the use of large foundational models for specific medical applications is gaining traction~\cite{medsam}. Here, the goal is to create large pre-trained models encompassing a large cohort of data with the potential to be fine-tuned to a number of relevant downstream tasks. However, pretraining at scale is challenging for medical imaging. In the case of OCT imaging, for instance, one of the largest public datasets contains roughly $100$K images from $4'600$ patients~\cite{kermany}. The recently published model, \textit{RETFound}~\cite{retfound} was trained on $736$K OCT images from $37$K, predominantly diabetic patients. Despite the unprecedented dataset scale, RETFound was evaluated by finetuning on only two tasks related to retinal pathology, wet AMD conversion prediction (treated as binary classification) using private data, and $5$-way disease classification on a small ($560$ images) public dataset \cite{octid}. 
This limits the generality of the conclusions that can be drawn about the utility of the learned representations for other OCT-related tasks such as biomarker detection or segmentation. Furthermore, while the use of finetuning evaluates the effectiveness of the pretrained model as initialization for training on a specific task, it does not reveal its capacity as a general-purpose frozen feature extractor for different OCT analysis tasks.

We develop a general-purpose model for OCT understanding and leverage self-supervised pretraining with a large and diverse OCT dataset. Our model is trained on a comparable number of OCT images, namely $698$K, but from a more diverse set of $41$K patients, representing a wider variety of retinal pathologies. This results in effective representations for downstream tasks even without full finetuning. We explore $6$ downstream tasks ($4$ on public datasets) comprising disease classification, biomarker detection, disease stage prediction, and fluid/layer segmentation. We benchmark performance for finetuning and linear probing following standard practice from the natural image domain \cite{mae,dinov2}. The former demonstrates the ability of our pretrained model to yield strong task-specific performance, while the latter reveals the generality of the pretrained but frozen representation. We are the first to explore the effect of scaling model and pretraining dataset size. 

While diagnosis using only OCT is the gold standard in many cases~\cite{oct_review}, retinal pathology screening typically involves more imaging modalities \cite{multimodal_retinal_imaging}. Several works have proposed network architectures that fuse OCT with other visual modalities, such as RGB or Infrared fundus images, for tasks such as disease classification \cite{multimodal_retinal,multimodal_dr_OCTA_IR,multimodal_dr_glaucoma,multimodal_amd} or atrophy and blood vessel segmentation \cite{multimodal_seg_a,multimodal_seg_b}. 
A second contribution of our work is to propose an extension of masked autoencoders \cite{mae} for self-supervised pretraining using paired visual modalities. Specifically, we employ paired OCT and Scanning Laser Ophthalmoscope Infrared images (referred to as IR). The pretrained model is then evaluated on multimodal (OCT $\&$ IR) and unimodal (omitting either modality) diabetic retinopathy stage prediction. To the best of our knowledge, this constitutes the first result supporting large-scale, multimodal self-supervised pretraining for enhancing performance in a downstream diagnostic task. 

\section{Method}
\label{sec:method}

Our method aims to extend unsupervised model pre-training to multiple modalities for improved representation learning. We first summarize the masked image modeling method~\cite{mae}, used for unimodal pre-training, and then describe our proposed method which extends it to multiple modalities. Both approaches are illustrated in Fig.~\ref{fig:fig2}.

\subsection{Preliminaries}
\label{subsec:prelim}

Our model is a vision transformer (ViT)~$f:\left(\real^D\right)^*\to \left(\real^D\right)^*$ that receives a finite sequence~$\s\in \left(\real^D\right)^*$ of $D$-dimensional vectors, typically known as \emph{tokens}, and produces a descriptor~$f(\s)$ of the same size as the input sequence. When working with images, visual transformers require that they be represented as a sequence of vectors. To this end, each image~$\x$ is split into a sequence~$(\x_1, \ldots,\x_L)$ of non-overlapping $p\times p$ patches and each patch is mapped into a $D$-dimensional embedding space, $\s_i = E(\x_i) + \mathbf{p}_i$, where $E$~is a learnable linear map and $\mathbf{p}_i$ is the standard sinusoidal positional embedding~\cite{attention_is_all}. The resulting sequence~$\s=(\s_1, \ldots, \s_L)$ is then fed as input to the ViT.

\subsection{Unimodal masked image modelling}

\label{subsec:mim}

Representation learning commonly involves pre-training ViT models using unsupervised tasks. In \cite{mae}, this task is masked image reconstruction, where the model is trained to reconstruct an whole image when only a random subset of its patches are given as input. Formally, the image sequence~$\s$ is first filtered to produce the sequence of \emph{visible patches}~$\bar{\s}$, where each token~$\s_i$ of the original sequence is dropped according to the value of a random Bernoulli variable~$m_i\sim Be(\rho)$. The \emph{masking ratio}~$\rho$ controls the proportion of dropped patches. The visible sequence is then fed to the ViT, producing a sequence of descriptors~$f(\bar{\s})$. To reconstruct the image from the descriptors, the descriptor sequence~$f(\bar{\s})$ is filled by inserting a learnable token with positional information, $\textit{<MASK>} + \mathbf{p}_i$, at the locations~$i$ corresponding to the dropped patches. This filled sequence is processed by a lightweight decoder, also modeled as a ViT, to produce the estimated reconstruction of the original image~$\hat{\x}$. The entire system is trained end-to-end minimizing the reconstruction error for the dropped patches,

\begin{equation}
    \mathcal{L}_\textrm{unimodal}(\x, \hat{\x}, \m) = \dfrac{1}{\sum_i m_i} \sum_i m_i \norm{\x_i - \hat{\x}_i}^2_2.
\end{equation}
After training, the reconstruction decoder is discarded.

\begin{figure}[t]
\centering
\includegraphics[width=0.95\linewidth]{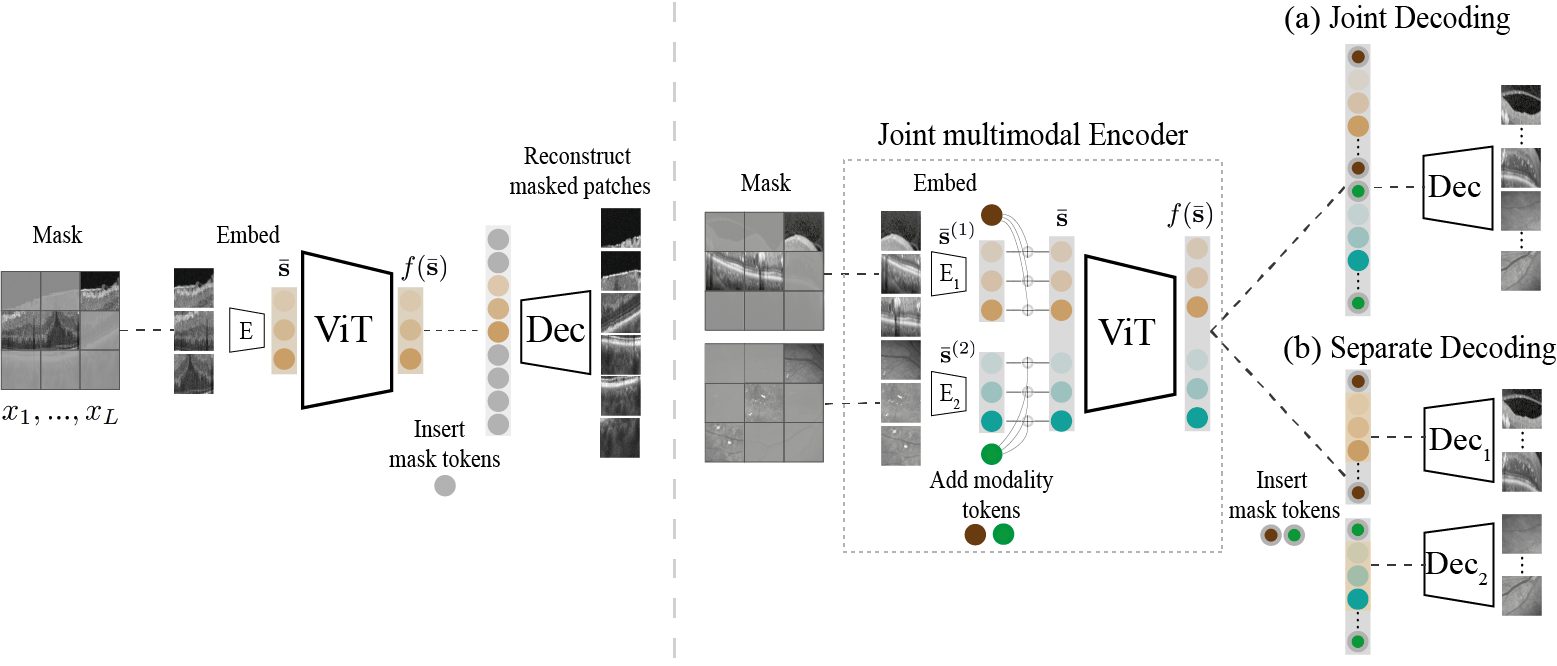}
\caption[]{Unimodal \cite{mae} and our proposed multimodal masked image modelling.}
\label{fig:fig2}
\end{figure}

\subsection{Multimodal masked image modelling}
\label{subsec:mmim}

Training the ViT~$f$ with multiple modalities enables it to learn shared representations across different image types, enhancing its ability to understand complex relationships. This results in more comprehensive and robust representations compared to using separate transformers for each modality. To this end, we extend the masked reconstruction task from~\cite{mae} to pre-train a transformer with paired multimodal images.

In particular, given two paired images~$(\x^{(1)}, \x^{(2)})$ coming from two modalities, we first compute their corresponding sequences of tokens~$\s^{(1)}$ and~$\s^{(2)}$ as explained in the previous section. The sequences are filtered with masks~$\m^{(1)}$ and~$\m^{(2)}$, resulting in the visible sequences~$\bar{\s}^{(1)}$ and $\bar{\s}^{(2)}$. The modality-wise mask ratios~$\rho^{(1)}$ and~$\rho^{(2)}$ used in this process are hyperparameters of our method. In order to allow~$f$ to differentiate tokens coming from different modalities, we sum the learnable tokens \emph{<MOD1>} and \emph{<MOD2>} to the elements of~$\bar{\s}^{(1)}$ and~$\bar{\s}^{(2)}$, respectively. The resulting sequences are concatenated into a joint sequence~$\bar{\s}$ and fed to~$f$ to produce the joint descriptors~$f(\bar{\s})$. These joint descriptors are split back per modality, and learnable tokens with positional information, $\textit{<MASK1>} + \mathbf{p}_i$ and $\textit{<MASK2>} + \mathbf{p}_i$, are inserted at the locations~$i$ of the masked tokens for each modality. Finally, the filled sequences are passed to the reconstruction model to generate the reconstructed images~$\hat{\x}^{(1)}$ and~$\hat{\x}^{(2)}$.

We consider two variants for the image reconstruction approach. The first variant (Fig.~\ref{fig:fig2}.a) uses a single decoder for both modalities, $D_\textrm{joint}$, except for the two final linear prediction layers that reconstruct pixels for each modality. 
The second variant (Fig.~\ref{fig:fig2}.b) uses separate decoders $D_1$ and~$D_2$ for each modality. This variant prevents any cross-modal interaction in the decoding layers. Instead, cross-modal interaction is limited to the attention layers of the encoder~$f$.

Regardless of the decoding approach, the multimodal reconstruction loss is 
\begin{equation}
    \mathcal{L}(\x^{(1)}, \x^{(2)}, \hat{\x}^{(1)}, \hat{\x}^{(2)}, \m^{(1)}, \m^{(2)}) = \sum_{n \in \{1,2\}} \mathcal{L}_\textrm{unimodal}(\x^{(n)}, \hat{\x}^{(n)}, \m^{(n)}).
\end{equation}
Once trained, our joint multimodal encoder can be used for both unimodal and multimodal tasks.

\begin{figure*}[t]
\centering
\includegraphics[width=0.99\linewidth]{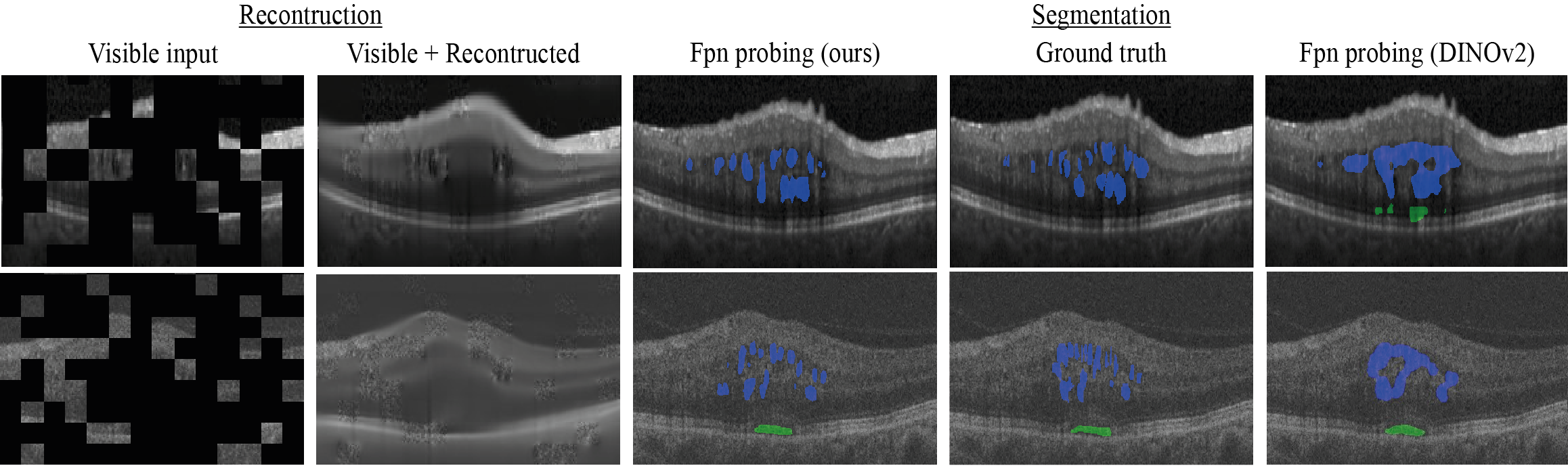}
    \caption[]{Examples of reconstructions by our model on unseen images from RETOUCH. The model's encoder, without any finetuning, combined with a lightweight feature pyramid network, produces fluid segmentations of higher quality than DINOv2 with the same approach.}
\label{fig:fig4}
\end{figure*}

\section{Experimental setup}

\subsubsection{Datasets and downstream tasks:}

For self-supervised pretraining we utilize an internally collected dataset of $589$K $2$D OCT images and $371$K IR fundus images from $36$K patients, with an average age of $58.4\pm 26.5$ and a variety of early and late AMD and DR patients, as well as Geopgraphic Atrophy, Retinal Vein Occlusion and Healthy patients. All data was collected during routine clinical practice at a hospital's ophthalmology clinic. 

For unimodal OCT pretraining, we combine our dataset with $109$K OCT images from $4686$ patients from the public dataset of \cite{kermany}, resulting in a total of $698$K OCT images from $41$K patients. We denote this dataset as \textbf{``Ours''}. For multimodal pretraining we use paired OCT and IR images. Pairs are formed using images captured from the same eye of the patient during the same hospital visit. In total, we use $592$K pairs from $371$K IR and $589$K OCT images. We denote this dataset as \textbf{``Our-m''}.

We evaluate our pretrained models on $6$ OCT analysis tasks and $1$ multimodal tasks involving both OCT and IR. Here, we briefly describe each task and provide more information in the supplementary section: \textbf{OCTID} (Cirrus) \cite{octid} is a disease classification dataset, of $572$ patients split into $5$ classes. %
\textbf{OCTDL} (Topcon) \cite{octdl} is a disease classification dataset that has $1618$ images from $631$ subjects split into $7$ classes. 
\textbf{Retouch} (Spectralis, Topcon, Cirrus) \cite{retouch} is a fluid segmentation dataset of $6936$ images from $71$ subjects with $3$ annotated classes. \textbf{AROI} (Cirrus) \cite{aroi} is a fluid and retinal layer segmentation dataset of $1136$ images from $23$ subjects with $3$ fluid and $3$ layer types annotated. \textbf{OctBiom} (Spectralis) is a biomarker detection (multi-label classification) dataset of $24$K images from $348$ patients with $9$ annotated biomarkers. \textbf{DRS} (Spectralis) is a dataset for disease stage classification of $1300$ pairs of OCT and IR images from $249$ patients with either non-proliferative or proliferative DR. 

\subsubsection{Implementation details:}

For pretraining, we use a batch size of $1024$ and train for $400$~epochs with masking ratios of $85\%$ for OCT and $65\%$ for IR. We initialize pretraining with the Imagenet self-supervised weights of~\cite{mae}. We ablate this choice in the supplementary material. We use linear learning rate warm-up followed by cosine decay with a peak value of $10^{-4}$. 

\subsubsection{Downstream task adaptation:}
For downstream tasks, we describe training hyper-parameters in the supplementary.
For classification and detection tasks, we input the spatially average-pooled features from the last layer of our model to a linear layer ($\sim20$K params) that produces $N_c$-dimensional logits ($N_c$ being the number of classes). For multimodal classification on DRS, we perform average pooling over each modality's tokens separately and concatenate the results. For segmentation, we append the lightweight feature pyramid (fpn) ($\sim8$M params) of \cite{vitd} to our model and train using the Lovasz softmax loss\cite{lovasz}. We provide results with ViT-B ($86$M) and ViT-L ($307$M) \cite{vit}.

\subsubsection{Baselines:}
We evaluate our method against three different existing methods:
\begin{itemize}
    \item \textbf{ImageNet} self-supervised pretraining (denoted as ``IN''), being widely adopted as the standard for transfer learning despite the domain gap between Imagenet and medical image domains \cite{matsoukas2023pretrained}.
    \item \textbf{RETFound} \cite{retfound} a ViT-L model trained with comparable amounts of OCT data.
    \item \textbf{DINOv2} \cite{dinov2} ViT-L and ViT-G ($1.1$B) which are widely adopted as frozen general-purpose feature extractors for computer vision tasks \cite{foundational_on_medical}.
\end{itemize}

\subsubsection{Experiments:}

We evaluate two settings: fine-tuning, wherein the whole model is updated, and linear/fpn probing, wherein only the additional parameters are trained. 

For all experiments, we repeat each run with $3$ different random seeds and report the mean and standard deviation for each metric. We select Macro F1 score (MF1) for multiclass and multi-label classification tasks, ROC-AUC for binary disease stage classification and mean intersection over union (mIoU) for fluid and layer segmentation. We show a representative metric per dataset and provide more metrics in the supplementary.

\section{Results}

\subsection{Unimodal OCT Model}

\begin{table*}[t]
\centering
\resizebox{.99\linewidth}{!}{
\begin{tabular}{ccccccccccc}
\toprule
\multirow{2}{*}\textbf{Model} & \multirow{2}{*}\textbf{Size} &
\multirow{2}{*}\textbf{Init} & \textbf{OCTID}   & \textbf{OCTDL} & \textbf{OctBiom} & \textbf{DR} & \textbf{Retouch} & \multicolumn{2}{c}{\textbf{AROI}}\\
&&& MF1  & MF1 & MF1 & ROC-AUC & mIoU (F) & mIoU (L) & mIoU (F)\\

\midrule

ViT-B & 86M & IN & 75.4\std{0.4} & 75.0\std{2.4} & 38.1\std{1.1} & 82.8\std{0.6} & 53.0\std{0.5} & 82.3\std{0.2}&  45.4\std{0.2}  \\ 
ViT-L & 303M & IN & 85.3\std{3.8} & 76.1\std{0.6} & 42.8\std{2.9} & 83.9\std{0.8} & 51.5\std{0.5} & 83.0\std{0.1} & 45.9\std{0.4} \\ 

DINOv2-L & 303M & LVD-142M & 86.3\std{1.2} & 74.7\std{0.1} & 51.3\std{1.4} & 79.9\std{2.0} & 54.9\std{0.2} & 80.0\std{0.1} & 47.2 \std{0.1} \\

DINOv2-G & 1.1B & LVD-142M & 89.5\std{0.7} & 73.7\std{2.3} &  50.7\std{0.4} & 79.0\std{0.8} & 58.6\std{0.3}  & 81.6\std{0.0}&  49.5\std{0.8} \\

\midrule

ViT-L & 303M & RETFound  & 85.3\std{0.6} & 82.7\std{0.7} & 39.8\std{0.4} & 83.3\std{0.1} & \textbf{61.0}\std{0.4} & \textbf{84.0}\std{0.8} & 
 \textbf{54.0}\std{0.7}\\

\midrule

ViT-B & 86M & Ours & \textbf{93.0}\std{0.1} & \un{83.6}\std{2.0} & \un{54.7}\std{0.4} & \un{89.1}\std{2.5} & 60.2\std{0.1} & 82.3\std{0.2} & 45.5\std{0.5} \\

ViT-L & 303M & Ours~$20\%$ & 84.4\std{0.6} & 83.0\std{0.7} & 49.1\std{0.4} & 88.1\std{0.1} & 59.6\std{0.4} & 83.3\std{0.0}&  45.5\std{0.4}  \\

\rowcolor{Gray3}

ViT-L & 303M & Ours  & \un{90.3}\std{0.6} & \textbf{84.6}\std{0.7} & \textbf{55.0}\std{0.4} & \textbf{90.9}\std{0.1} & \un{60.5}\std{0.4} & \textbf{84.0}\std{0.1} & 
 \un{53.5}\std{0.1}\\

\bottomrule
\end{tabular}}
\quad
\quad
\caption{Linear/fpn probing comparisons. \textbf{Bold} indicates best per dataset, \un{underline} indicates $\nth{2}$ best. mIoU (F) and (L) correspond to mIoU averaged from fluid and retinal layer classes, respectively. ``Ours $20\%$'' refers to pretraining on $20\%$ of our complete dataset.}
\label{tab:lp_all}
\end{table*}

\noindent\textbf{Linear/fpn Probing:} We consistently outperform DINOv2, which was trained $200\times$ more data and with a combination of masked image modelling and self-distillation (Table~\ref{tab:lp_all}). This holds for equal model sizes ($303$M) and ViT-G ($1.1$B), hinting that large vision models from the natural image domain are competitive but sub-optimal as frozen feature extractors for OCT understanding tasks. We observe that pretraining on $20\%$ (``ours $20\%$'') of our dataset decreases performance on all tasks, showcasing the benefit of scaling the dataset size. We also outperform RETFound on $4/6$ tasks while marginally outperforming for fluid segmentation by $0.5\%$ mIoU on RETOUCH and $0.5\%$ mIoU on AROI. Overall, our model achieves the most consistent performance across all tasks. Notably, despite having been pretrained only using Spectralis images, it remains effective across device types, despite not being finetuning.

\vspace{2mm}

\begin{table*}[t]
\centering
\resizebox{.99\linewidth}{!}{
\begin{tabular}{ccccccccccc}
\toprule
\multirow{2}{*}\textbf{Model} & \multirow{2}{*}\textbf{Size} &
\multirow{2}{*}\textbf{Init} & \textbf{OCTID}   & \textbf{OCTDL} & \textbf{OctBiom} & \textbf{DRS} & \textbf{Retouch} & \multicolumn{2}{c}{\textbf{AROI}}\\
&&               & MF1  & MF1 & MF1 & ROC-AUC & mIoU (F) & mIoU (L) & mIoU (F)\\

\midrule

ViT-B & 86M & IN & 75.4\std{0.4} & 88.1\std{0.5} & 70.6\std{0.7} &  91.4\std{0.3} & 67.4\std{0.4} &  86.5\std{0.1}&  59.1\std{0.5} \\ 
ViT-L & 303M & IN & 85.3\std{3.8} & 89.0\std{2.1} & 70.8\std{0.8} &  90.6\std{0.5} & 67.4\std{0.5} &  87.5\std{0.0} &  63.7\std{0.1} \\

\midrule

ViT-L & 303M & RETFound  & \un{94.5}\std{2.0} & \un{90.3}\std{1.7} & 68.6\std{2.1} & \un{91.2}\std{0.2} & \un{68.3}\std{0.2} & 87.5\std{0.0} & 62.5\std{0.4} \\

\midrule

ViT-B & 86M & Ours & 93.8\std{0.1} & \textbf{90.8}\std{1.2} & 71.5\std{1.1} &  91.5\std{0.3} & 67.5\std{0.3} & 87.3\std{0.0} & 61.7\std{0.5} \\

ViT-L & 303M & Ours~$20\%$ & \un{94.5}\std{1.0} & 88.9\std{1.4}& \un{73.3}\std{0.4}  & 92.0\std{0.4} & \un{68.4}\std{0.2} & 87.4\std{0.2}&  63.3\std{0.2}    \\
\rowcolor{Gray3}

ViT-L & 303M & Ours  & \textbf{94.7}\std{0.7}  & 89.9\std{1.2} & \textbf{74.7}\std{0.6} & \textbf{92.4}\std{0.4} & \textbf{69.0}\std{0.0} & \textbf{87.8}\std{0.0} & \textbf{64.0}\std{0.1}  \\

\bottomrule
\end{tabular}}
\quad
\quad
\caption{Finetuning comparisons. \textbf{Bold} indicates best per dataset, \un{underline} indicates $\nth{2}$ best. mIoU (F) and (L) correspond to mIoU averaged from fluid and retinal layer classes, respectively.}
\label{tab:ft_all)}
\end{table*}

\noindent\textbf{Finetuning:} We outperform Imagenet initialization on all tasks (Table \ref{tab:ft_all)}). Relative to RETFound, our model achieves significantly higher performance for fluid segmentation (RETOUCH, AROI), biomarker detection (OctBiom) and disease stage classification (DRS). For disease classification (OCTID, OCTDL), and layer segmentation (AROI) the difference between RETFound and our model is marginal ($0.2-0.4\%$). We also observe that increasing model size improves performance for fluid segmentation and biomarker detection while for classification tasks the differences are marginal. 

\subsection{Multimodal model}
We evaluate our multimodal model, pretrained using the method of Section~\ref{subsec:mmim}, on \textbf{DRS} (Tab.~\ref{tab:DR}), for $3$ types of input: OCT $\&$ IR, OCT, IR. We discuss differences in terms of ROC-AUC but similar conclusions can be drawn for PR-AUC.

\noindent\textbf{OCT $\&$ IR:} Our best multimodal model improves over our best OCT-only baseline ($+2.6\%$, for finetuning and $+2.8\%$ for linear probing). This demonstrates that multimodality is beneficial for this task. 
We find that using IN pretraining for the joint encoder (MViT) is significantly outperformed by our multimodal pretraining ($+7\%$ for finetuning and $+5.8\%$ for linear probing). 

\noindent\textbf{OCT:} Our model performs worse than IN ($-2.3\%$) as well as all other models trained only on OCT. However, it outperforms the former ($+6.4\%$) and performs on par with the latter, with linear probing. 

\noindent\textbf{IR:} Our model outperforms IN with both finetuning ($+4.7\%$) and linear probing ($+6.5\%$). Overall these result showcase, that our multimodal model is able to remain effective even with missing modalities. 

\noindent\textbf{Decoding:} We find that the two decoding approaches discussed in Section~\ref{subsec:mmim}, joint and separate decoding, perform comparably with finetuning, but the latter achieves the highest performance with linear probing ($+2\%$).  

\begin{table*}[t]
\centering
\resizebox{.95\linewidth}{!}{
\begin{tabular}{l*{10}c} 
\toprule 
\multirow{2}{*}{\textbf{Model}} &
\multirow{2}{*}{\textbf{Init}} & \multicolumn{2}{c}{\textbf{Modalities}} &
\multicolumn{3}{c}{\textbf{Finetuning}} & \multicolumn{3}{c}{\textbf{Linear Probing}}\\
\cmidrule{3-4}\cmidrule{6-7} \cmidrule{9-10} 
&& OCT & IR & & ROC-AUC & PR-AUC & & ROC-AUC &  PR-AUC \\

\midrule
ViT-L  & IN &  \tick  &       && 90.6\std{0.5} & 94.2\std{0.4} &&  83.9\std{0.8} & 90.1\std{0.4} \\

ViT-L  & IN-RETFound &  \tick & && 91.2\std{0.2} & 94.4\std{0.1} &&  83.3\std{2.9} & 89.4\std{2.4} \\    

ViT-B  & Ours & \tick & && 91.5\std{0.3} & 95.5\std{0.1} && 89.1\std{2.5} & 93.5\std{2.2} \\
\rowcolor{Gray3}

ViT-L  & Ours & \tick & && \textbf{92.4}\std{0.4} & \textbf{96.0}\std{0.3} && \textbf{90.9}\std{0.1} & 94.8\std{0.2}  \\

MViT-B  & Ours-m-$D_{joint}$ & \tick &  && 88.3\std{0.6} & 93.8\std{0.4}&& 90.3\std{0.7} & \textbf{95.1}\std{0.2}  \\

\midrule

ViT-L  & IN &         & \tick && 79.5\std{1.3} & 88.0\std{0.9} && 83.3\std{0.5} & 89.4\std{0.2}  \\    
\rowcolor{Gray3}

MViT-B  & Ours-m-$D_{joint}$ &&    \tick && \textbf{84.2}\std{0.8} & \textbf{90.3}\std{1.2}  && \textbf{89.8}\std{0.4} & \textbf{94.1}\std{0.2}    \\

\midrule

MViT-B  & IN &  \tick & \tick && 87.9\std{2.5} & 92.7\std{2.1} && 87.9\std{1.2} & 92.9\std{0.7}  \\  

\rowcolor{Gray3}
MViT-B  & Ours-m-$D_{joint}$ & \tick & \tick && \textbf{95.0}\std{0.3} & \textbf{97.3}\std{0.2} &&  91.7\std{0.2} & 95.4\std{0.1} \\
\rowcolor{Gray3}
MViT-B  & Ours-m-$D_{sep}$& \tick & \tick && 94.1\std{1.7} & 96.7\std{0.9} && \textbf{93.7}\std{0.7} & \textbf{96.6}\std{0.1}\\

\bottomrule
\end{tabular}}

\caption{Diabetic retinopathy stage classification. \textbf{Bold} indicates best per input type (OCT $\&$ IR, OCT, IR).}
\label{tab:DR}
\end{table*}

\section{Conclusion}
We presented the first extensive evaluation of the effectiveness of masked image modelling for learning versatile representations for OCT understanding. Our model, which will be published, is trained on diverse patient data and outperforms large vision models from the natural image domain as well as competing models trained on OCT data. Furthermore, we proposed a self-supervised method for pretraining using multiple retinal imaging modalities. We show multimodal models pretrained with our method achieve strong performance on a multimodal diagnostic task and can also remain effective with missing modalities. Overall, our findings advocate for further investigation into the roles of large-scale pretraining and multimodality in retinal image analysis.

\bibliographystyle{splncs04}
\bibliography{main.bib}

\section{Supplementary Material}
\begin{figure*}[ht]
\centering
\includegraphics[width=0.99\textwidth]{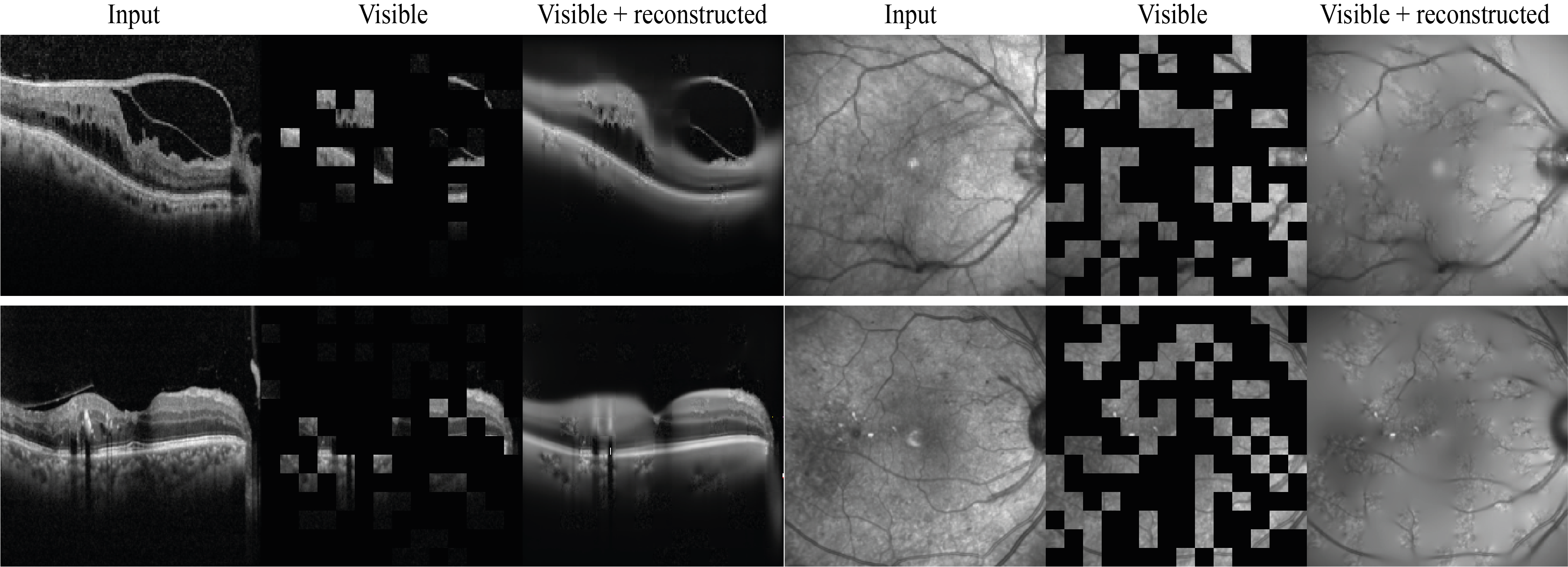}
\caption[]{Multimodal reconstruction examples. OCT and IR images are encoded by our joint multimodal encoder and decoded by modality specific decoders.}
\label{fig:fig1_supp}
\end{figure*}
\begin{table}
\resizebox{0.99\linewidth}{!}{
    \centering
    \begin{tabular}{ccccccc}
    \toprule
     Dataset & Subjects per split & Image per split & Modalities & Scanner & Pathology & Task  \\
     \midrule
     OctBiom & 288/20/40 &  23K/1K/1K  & OCT & Spectralis & DR (w/wo DME ), AMD (All stages)  &Biomarker Detection \\
     \midrule
     DR & 149/30/70 & 972/111/317 & OCT, IR & Spectralis & DR (P and NP)  &cls (PDR vs NPDR) \\
     \midrule
     \midrule
     RETOUCH & 52/19  & 5106/1830 & OCT & Spectralis/Cirrus/Topcon & AMD and RVO &  Fluid Segm. \\
     \midrule
     
     AROI  & 13/13  & 596/509 &OCT & Cirrus & AMD & Fluid $\&$ Layer Segm. \\
     \midrule
     OCTID  & N/A  & 316/82/174 & OCT & Cirrus & AMD, DR, MacHole, CSR & Classification  \\
    \midrule
     OCTDL  & 410/64/157  & 903/360/355 & OCT& Topcon & AMD, DME, ERM, RAO, RVO, VID & Classification   \\

     \bottomrule
    \end{tabular}}
    \quad
    \caption{: We present details for all the datasets used for finetuning and linear probing experiments. We split classification and detection datasets in train/val/test splits stratified according to pathology. For these datasets we use the best checkpoint on the validation set to compute metrics on the test set. For segmentation datasets we use two splits (train/val) and report performance of the final training checkpoint on val. To enable reproduction of our results on public datasets we will release all dataset splitting information in our code repository.}
    \label{tab:datasets}
\end{table}
\begin{table*}[ht]
\centering
\resizebox{.99\linewidth}{!}{
\begin{tabular}{lcccccccccc}
\toprule
\multirow{2}{*}{\textbf{Model}} &
\multirow{2}{*}{\textbf{Init}} &
\multicolumn{4}{c}{\textbf{Finetuning}} & \multicolumn{5}{c}{\textbf{Linear Probing}}\\
\cmidrule{3-6}\cmidrule{8-11}
&&  \mc{ROC-AUC} & \mc{MF1} & \mc{MP} & \mc{Acc}  &  & \mc{ROC-AUC} & \mc{MF1} & \mc{MP} & \mc{Acc}\\

ViT-L  & IN   & 99.4\std{0.2} & 91.3\std{0.5} & 91.4\std{0.2} & 93.3\std{0.2} & & 99.0\std{0.1} & 85.3\std{3.8} & 87.6\std{2.8} & 88.5\std{2.8}   \\
ViT-L & RETFound  & \textbf{99.7}\std{1.5} & 94.5\std{2.0} & 94.6\std{2.0} & 95.7\std{1.5} &   & 99.3\std{0.0} & 85.3\std{1.4} & 86.9\std{1.0} & 88.5\std{1.2} \\  

\midrule

ViT-B  & Ours & 99.3\std{0.1} & 93.8\std{0.6} & 94.2\std{0.9} & 95.4\std{0.4} &    & \textbf{99.6}\std{0.0} & \textbf{93.0}\std{0.1}  & \textbf{93.8} \std{0.9} & \textbf{94.4}\std{0.7} \\

ViT-L  & Ours $20\%$  & 99.7\std{0.0} & 94.5\std{1.0} & 94.7\std{1.1}  & 95.8\std{0.7}  & & 98.8\std{0.1} & 84.4\std{3.7} & 85.6\std{2.8} & 87.3\std{2.9} \\

\rowcolor{Gray3}

ViT-L  & Ours & 99.6\std{0.4} & \textbf{94.7}\std{0.7} & \textbf{94.9}\std{0.8} & \textbf{95.8}\std{0.5}  &&  99.5\std{0.0} & 90.3\std{0.6} & 90.5 \std{0.7} & 92.1\std{0.5} \\

\midrule
\midrule 

ViT-L  & IN   & 98.9\std{0.2} & 89.0\std{2.1} & \un{90.8}\std{1.2} & 92.9\std{0.8} & &  96.8\std{0.2} & 76.1\std{0.6} & 75.3\std{2.4} & 83.2\std{0.6} \\

ViT-L & RETFound & \textbf{99.2}\std{0.3} & \un{90.3}\std{1.7} & 88.7\std{3.1} & \textbf{93.5}\std{0.7} & & \un{98.4}\std{0.2} & \un{82.7}\std{2.0} & \un{81.4}\std{1.4} & \un{86.9}\std{1.0}\\

\hline

ViT-B  & Ours & 99.1\std{0.1} & \textbf{90.8}\std{1.2} & \textbf{89.8}\std{1.2} & 89.7\std{1.3} & & 98.5\std{0.2}  & 83.6\std{2.0} & 81.4\std{3.0} & \textbf{89.6}\std{0.3}\\

ViT-L  & Ours $20\%$ & 99.0\std{0.1} & 88.9\std{1.4} & 87.9\std{2.3} & 91.9\std{1.0} & & 98.5\std{0.0} & 83.0\std{0.7} & 81.8\std{1.8} & 89.5\std{0.2} \\

\rowcolor{Gray3}
ViT-L  & Ours & \textbf{99.2}\std{0.2} & 89.9\std{1.2} & 89.3\std{1.6} & \un{92.9}\std{0.3} & &  \textbf{98.6}\std{0.1} & \textbf{84.6}\std{0.7} & \textbf{81.8}\std{0.7} & 89.3\std{0.1} \\

\bottomrule
\end{tabular}}
\caption{Additional metrics for OCTID (top $4$ rows) and OCTDL (bottom $4$ rows)}
\label{tab:OCTDL}
\end{table*}
\begin{table*}[ht]
\centering
\resizebox{.99\linewidth}{!}{
\begin{tabular}{lcccccccccccc}
\toprule
\multirow{2}{*}{\textbf{Model}} &
\multirow{2}{*}{\textbf{Init}} &
\multicolumn{5}{c}{\textbf{Finetuning}} & \multicolumn{6}{c}{\textbf{Linear Probing}}\\
\cmidrule{3-7}\cmidrule{9-13}
&&  \mc{ROC-AUC} & \mc{MF1} & \mc{mF1} & \mc{MAP} & \mc{mAP} &  & \mc{ROC-AUC} & \mc{MF1} & \mc{mF1} & \mc{MAP} & \mc{mAP}\\
\midrule

\mc{ViT-L}  & \mc{IN}   & \un{94.6}\std{0.3} & \un{70.8}\std{0.8} & \un{74.7}\std{0.7} & 82.9\std{2.0} & 83.8\std{1.0} &  &  \un{83.1}\std{1.4} & \un{42.8}\std{2.9} & \un{51.2}\std{1.8} & 54.3\std{0.8} & \un{58.2}\std{1.2}\\

\mc{ViT-L}  & \mc{RETFound} & 94.3\std{0.2} & 68.6\std{2.1} & 73.1\std{1.1} & \textbf{84.3}\std{0.1} & \un{85.6}\std{0.7}  & & 78.0\std{0.9} & 39.8\std{1.6} & 47.3\std{1.4} & \un{56.2}\std{0.8} & 50.1\std{1.4}\\  

\hline

\mc{ViT-B}  & \mc{Ours}   & 94.3\std{0.2} & 71.5\std{1.1} & 74.5\std{0.3} & 82.6\std{0.6} & 84.3\std{0.4} & & 85.7\std{0.4} & \textbf{55.0}\std{0.4} & \textbf{59.7}\std{0.2} & 63.6\std{0.2} & \textbf{65.5}\std{0.3}\\

\mc{ViT-L}  & \mc{Ours $20\%$}   & 94.6\std{0.1} & 73.3\std{0.4} & 75.1\std{0.4} & 84.7\std{0.3} & 85.2\std{0.2} &  & 83.4\std{1.2} & 49.1\std{2.0} & 54.2\std{1.8} & 62.8\std{0.4} & 59.1\std{1.8}\\

\rowcolor{Gray3}
\mc{ViT-L}  & \mc{Ours}   & \textbf{95.1}\std{0.1} & \textbf{74.7}\std{0.6} & \textbf{76.1}\std{0.6} & \un{83.9}\std{0.3} & \textbf{85.7}\std{0.4} & & \textbf{86.6}\std{0.3} & 54.7\std{1.0} & 57.3\std{0.6} & \textbf{66.9}\std{0.4} & 62.8\std{0.2} \\

\bottomrule
\end{tabular}}
\caption{OctBiom}
\label{tab:octbiom}
\end{table*}
\begin{table*}[h]
\centering
\resizebox{.99\linewidth}{!}{
\begin{tabular}{lccccccccccc}
\toprule
\multirow{2}{*}{\textbf{Model}} &
\multirow{2}{*}{\textbf{Init}} &

\multicolumn{4}{c}{\textbf{Finetuning}} & \multicolumn{5}{c}{\textbf{FPN Probing}}\\
\cmidrule{3-6} \cmidrule{8-11} 
&& mIoU & IRF & SRF & PED&  & mIoU & IRF & SRF & PED\\
\midrule

ViTD-L  & \mc{IN}   & 67.4\std{0.5} & 57.0\std{0.1} & 80.1\std{0.6} & 65.1\std{1.1} &  &  51.5\std{0.5} & 51.1\std{0.0} & 67.4\std{0.5} & 35.8\std{2.0}\\
ViTD-L  & RETFound  & \un{68.3}\std{0.2} & \textbf{58.3}\std{0.0} & 81.3\std{0.1} & 65.1\std{0.5} &  & \textbf{61.0}\std{0.3} & \un{52.6}\std{0.3} & \textbf{73.3}\std{0.4} & \textbf{57.0}\std{0.7} \\   

\hline

ViTD-B  & Ours  & 67.5\std{0.3} & 57.3\std{0.2} & 79.9\std{0.4} & 65.3\std{0.5} & & 60.2\std{0.1} & 53.3\std{0.2} & 72.4\std{0.4} & 55.0\std{0.2} \\

ViTD-L  & \mc{Ours $20\%$}  & 68.4\std{0.2} & 57.1\std{0.6}& 81.6\std{0.4}& 66.7\std{0.2} &  & 59.6\std{0.1} & 52.1\std{0.1} & \un{71.3}\std{0.5} & 55.4\std{0.8} \\

\rowcolor{Gray3}
ViTD-L  & Ours  & \textbf{69.0}\std{0.0} & \textbf{58.3}\std{0.1} & \textbf{82.3}\std{0.1} & \textbf{66.5}\std{0.2}  &  & \un{60.5}\std{0.4} & \textbf{53.9}\std{0.2} & 70.5\std{0.4} & \textbf{57.0}\std{1.3} \\

\bottomrule
\end{tabular}}
\caption{Retouch}
\label{tab:retouch}
\end{table*}
\begin{table*}[h]
\centering
\resizebox{.99\linewidth}{!}{
\begin{tabular}{lcccccccccc}
\toprule
\multirow{2}{*}{\textbf{Model}} &
\multirow{2}{*}{\textbf{Pretraining}} &
\multicolumn{2}{c}{\textbf{OctBiom} (MF1)} & \multicolumn{3}{c}{\textbf{RETOUCH} (mIoU)} & \multicolumn{3}{c}{\textbf{OCTID} (MF1)}\\
\cmidrule{3-4}\cmidrule{6-7}\cmidrule{9-10}
&&  Finetuning & Linear probing &  & Finetuning & FPN probing & & Finetuning & Linear probing \\

\midrule
ViT-B  & random $\rightarrow $ Ours  & 69.4\std{1.1} & 47.5\std{0.3} &  & 65.7\std{0.2} & 57.9\std{0.3} & & 93.5\std{0.9} & 90.0\std{1.2} \\
ViT-B  & IN   $\rightarrow $ Ours & 71.5\std{1.1} & 55.0\std{0.4} &  & 67.5\std{0.3} & 60.2\std{0.1}  &  & 93.8\std{0.6} & 93.0\std{0.1}  \\
\midrule
\rowcolor{Gray2}
& $\Delta$ & +2.1 & +7.5 & & +1.8 & +2.3 & & +0.3 & +3.0 \\ 

\bottomrule
\end{tabular}}
\caption{The effect of initializing pretraining from Imagenet.}
\label{tab:pretrain_init_IN)}
\end{table*}

\begin{table*}[htb]
  \centering
  
\resizebox{.99\linewidth}{!}{
\begin{tabular}{*{12}c}
\toprule
\tbf{Dataset} &\tbf{Mode} & \tbf{Init} & \textbf{Backbone} & \textbf{Res}   & \textbf{Peak lr}  & \textbf{Weight Decay} & \textbf{Stage Decay} & \textbf{Drop path} & \textbf{Epochs/Batch} & \textbf{optim} & \textbf{EMA} \\
\midrule

\multirow{2}{*}{OCTID} & {ft} & Ours/IN/RETF & \mc{ViT-B/L} & $(224,224)$ & $0.5*10^{-4}$  & $0.05$ & no & 0.3 &  $100/32$ & AdamW & no \\ 

 & lp &  Ours/IN/RETF/DINOv2 & \mc{ViT-B/L} &  $(224,224)$ & $10^{-2}$ & $0.05$ & N/A & N/A & \mc{$100/64$} & AdamW & no\\

\midrule

 \multirow{2}{*}{OCTDL}   & {ft} & Ours/IN/RETF & \mc{ViT-B/L} & $(224,224)$ & $0.5*10^{-4}$  & $0.2$ & no & 0.3 &  $100/32$ & AdamW & no\\ 

& lp &  Ours/IN/RETF/DINOv2  & \mc{ViT-B/L} &  $(224,224)$ & $10^{-2}$ & $0.05$ & N/A & N/A & \mc{$100/64$} & AdamW & no \\

\midrule

 \multirow{3}{*}{RETOUCH}   & {ft} & Ours/IN/RETF & \mc{ViT-B/L} & $(416,416)$ & $10^{-4}$  & $0.2$ & 0.9 (Ours, RETFound), no (IN) & 0.3 (B), 0.4 (L) &  $100/16$ & AdamW & no \\ 

& fpn-p &  Ours/IN/RETF  & \mc{ViT-B/L} &  $(416,416)$ & $10^{-3}$ & $0.05$ & N/A & N/A & \mc{$100/16$} & AdamW & no \\

& fpn-p &  DINOv2  & \mc{ViT/L} &  $(420,420)$ & $10^{-3}$ & $0.05$ & N/A & N/A & \mc{$100/16$} & AdamW & no \\

\midrule

 \multirow{3}{*}{AROI}   & {ft} & Ours/IN/RETF & \mc{ViT-B/L} & $(496,496)$ & $10^{-4}$  & $0.2$ & 0.9 (Ours, RETFound), no (IN) & 0.3  &  $200/16$ & AdamW & no \\ 

& fpn-p &  Ours/IN/RETF  & \mc{ViT-B/L} &  $(496,496)$ & $10^{-3}$ & $0.05$ & N/A & N/A & \mc{$200/16$} & AdamW & no \\

& fpn-p &  DINOv2  & \mc{ViT/L} &  $(504,504)$ & $10^{-3}$ & $0.05$ & N/A & N/A & \mc{$200/16$} & AdamW & no \\

\midrule

 \multirow{2}{*}{OctBiom}   & {ft} & Ours/IN/RETF & \mc{ViT-B/L} & $(416,416)$ & $10^{-3}$  & $0.1$ & 0.9 (Ours, RETFound) & 0.3  &  $100/16$ & AdamW &  (Ours, RETF) 0.999   \\ 

& lp &  Ours/IN/RETF/DINOv2  & \mc{ViT-B/L} &  $(416,416)$ & $10^{-2}$ & $0$ & N/A & N/A & \mc{$100/1024$} & SGD & no \\

\midrule

 \multirow{4}{*}{DR}   & {ft} & Ours/IN/RETF & \mc{ViT-B/L} & $(224,224)$ & $10^{-4}$  & $0.2$ & 0.5  & 0.3  &  $400/64$ & AdamW &   no   \\ 

& lp &  Ours/IN/RETF/DINOv2  & \mc{ViT-B/L} &  $(224,224)$ & $10^{-2}$ & $0$ & N/A & N/A & \mc{$400/128$} & SGD & no \\

& ft &  Ours-m/IN  & \mc{MViT-B} &  $(224,224)$ & $10^{-4}$ & $0.2$ & N/A & N/A & \mc{$400/64$} & ADAMW & no \\
& lp &  Ours-m/IN  & \mc{MViT-B} &  $(224,224)$ & $10^{-2}$ & $0$ & N/A & N/A & \mc{$400/128$} & SGD & no \\

\bottomrule
\end{tabular}
} 
\caption{\textbf{Training hyperparameters}. All runs use a learning rate warmup followed by cosine decay. By ``ft'', ``lp'', ``fpn-p'' we denote finetuning, linear probing and fpn probing, respectively. To reduce runtime, with higher resolution segmentation datasets (AROI, RETOUCH) we extract regions of interest using simple intensity thresholding. This  We then train and test on those ROIs. For all runs, we use flip, colorjitter and random scaling for data augmentation.}
\label{tab:hyperparams_downstream} 
\end{table*}

\end{document}